# Cognitive networks highlight differences and similarities in the STEM mindsets of human and LLM-simulated trainees, experts and academics

Edith Haim[1*], Lars van den Bergh[2*], Cynthia S. Q. Siew[3], Yoed N. Kenett, Daniele Marinazzo, Massimo Stella[1]

[1]*CogNosco Lab, Department of Psychology and Cognitive Science, University of Trento, Italy*
[2]*Department of Data Analysis, Ghent University, Belgium*
[3]*Department of Psychology, National University of Singapore*
* Co-first authors

**Author Note**

| | |
|---|---|
| Edith Haim | https://orcid.org/0009-0000-4244-904X |
| Lars van den Bergh | - |
| Cynthia S. Q. Siew | https://orcid.org/0000-0003-3384-7374 |
| Yoed N. Kenett | https://orcid.org/0000-0003-3872-7689 |
| Daniele Marinazzo | https://orcid.org/0000-0002-9803-0122 |
| Massimo Stella | https://orcid.org/0000-0003-1810-9699 |

We have no conflict of interest to disclose.
Correspondence concerning this article should be addressed to:
Edith Haim and Massimo Stella, CogNosco Lab, University of Trento, Corso Bettini 31, 38068 Rovereto TN, Italy, Emails: edith.haim@unitn.it, massimo.stella-1@unitn.it.

**Abstract**

Understanding attitudes towards STEM means quantifying the cognitive and emotional ways in which individuals, and potentially large language models too, conceptualise such subjects. This study uses behavioural forma mentis networks (BFMNs) to investigate the STEM-focused mindset, i.e. ways of associating and perceiving ideas, of 177 human participants and 177 artificial humans simulated by GPT-3.5. Participants were split in 3 groups - trainees, experts and academics - to compare the influence of expertise level on their mindsets. The results revealed that human forma mentis networks exhibited significantly higher clustering coefficients compared to GPT-3.5's, indicating that human mindsets displayed a tendency to form and close triads of conceptual associations while recollecting STEM ideas. Human experts, in particular, demonstrated robust clustering coefficients, reflecting better integration of STEM concepts into their cognitive networks. In contrast, GPT-3.5 produced sparser networks with weaker clustering, highlighting its limitations in replicating human-like mindsets. Furthermore, both human and GPT mindsets framed "mathematics" in neutral/positive terms, differently from STEM high-schoolers, researchers and other large language models sampled in other works. This research contributes to understanding how mindset structure can provide cognitive insights about memory structure and machine limitations.

*Keywords:* cognitive network science; mindset measurement; associative knowledge; artificial intelligence; simulated participants.

# Cognitive networks highlight differences and similarities in the STEM mindsets of human and LLM-simulated trainees, experts and academics

## 1. Introduction

Understanding how individuals form cognitive and emotional associations with STEM (science, technology, engineering, and mathematics) disciplines is essential for improving student engagement and addressing barriers to success in these fields (Abramski et al., 2023). Negative emotions, such as anxiety and frustration, are common in STEM education and have been shown to lead to avoidance behaviour, reduced academic achievement, and growing disinterest in pursuing STEM careers (Ashcraft, 2002; Pekrun, 2006). Given the critical role STEM plays in driving innovation and economic growth, these trends pose significant challenges to the development of a skilled workforce capable of addressing future societal needs (Luo et al., 2021). Stereotypical negative beliefs about STEM careers (e.g. that a career in STEM involves boring work in unpleasant, isolated surroundings) worsen this issue by negatively influencing students´ interest in STEM and their expectations regarding STEM-related professions. Negative outcome expectations, such as perceiving STEM careers as unachievable, unrewarding or incompatible with personal interests, reduce the appeal of STEM professions (Luo et al., 2021). Together, these factors create a feedback loop where stereotypes reinforce negative attitudes, discouraging students from entering STEM fields and worsening workforce shortages (Ashcraft, 2002; Pekrun, 2006).

Studies have shown that learners often develop fragmented or limited understanding of STEM topics, which can increase feelings of anxiety and disengagement (Haim et al., 2024a; Kubat & Guray, 2018; Vosniadou, 1994). Additionally, emotions are deeply intertwined with cognitive processes, shaping how individuals perceive, retain, and engage with STEM-related material (Immordino-Yang & Damasio, 2007). By exploring these associations, educators can develop more effective strategies to improve student engagement and learning outcomes in STEM disciplines (Abramski et al., 2023; Stella et al., 2023).

Recent research emphasises the importance of mapping the cognitive and emotional associations individuals form with STEM concepts to better understand the underlying factors that influence attitudes and behaviours in educational settings (Siew et al., 2019a; Stella et al., 2019). These cognitive and emotional patterns can be measured with behavioural forma mentis networks (BFMNs) which are a powerful tool for modelling associative memory via the proxy of memory recalls, i.e. continued free associations, and valence ratings. This combination of associations between ideas perceived in potentially different emotional ways can reflect the so-called mindset of either an individual (Stella, 2022a) or a group (Stella et al., 2019). Differently than for a textual forma mentis network that considers large samples of texts (Haim et al., 2025), in a behavioural forma mentis network associations are sampled from a continued free association task, where participants are prompted to respond to a stimulus word with the first three words that come to mind (De Deyne et al., 2013). This method provides insights into the associative links within an individual's mental lexicon. Free associations reflect the structure of semantic memory and are shaped by both experiences and linguistic factors (De Deyne et al., 2013). Thus, free associations offer a window into the cognitive processes underlying knowledge organisation (Wulff et al., 2022). Behavioural forma mentis networks (BFMNs) are constructed from these associations, where nodes represent concepts (stimulus words and their associations) and edges indicate the connections between them. In addition to semantic relationships, BFMNs also capture an emotional dimension through the incorporation of valence scores (Stella et al., 2019). By combining semantic associations with valence scores, BFMNs provide a comprehensive representation of both the cognitive and emotional aspects of an individual's mindset. These networks allow us to examine not only how concepts are linked but also how they are emotionally framed, which plays a crucial role in how attitudes and perceptions towards a given topic are shaped (Stella, 2022a).

Emotional valence, as defined also in Russel´s circumplex model of affect (1980), is a fundamental dimension of emotional experience, representing the intrinsic positive or negative dimension of a concept. Together with arousal (intensity of emotion), valence can form a basis of affective states (Russel, 1980). The BFMN approach focuses primarily on valence attributes (positive, negative or neutral) by reporting how words with different valences are linked with each other. Through a cross-validation with another dataset for valence and arousal, Stella and colleagues showed that in BFMNs, negative words linked with many other negative words tend to also have higher arousal compared to negative words linked with mostly neutral/positive words. This pattern indicates that in BFMNs, concepts rated as negative and linked with other negative concepts (e.g. "maths" in the mindset of high-schoolers) might indicate anxious perceptions (Stella et al., 2019).

Despite growing interest in this area, there is a significant gap in understanding how cognitive associations and emotional valences differ across various groups, such as trainees in comparison to experts in STEM fields. Previous research has often focused on student populations, neglecting the perspectives of more experienced individuals, such as experts working outside academia (Rice et al., 2013; Shin et al., 2016).

Furthermore, the potential biases and limitations of AI models, like GPT-3.5, in replicating human emotional and cognitive patterns have not been thoroughly explored (Abramski et al., 2023). The rise of artificial intelligence (AI) offers new opportunities to study cognitive and emotional patterns in education. AI-driven models, such as large language models (LLMs), offer novel ways to simulate and study human cognitive processes, providing insights that are difficult to obtain through traditional psychological approaches (LeCun et al., 2015; Abramski et al., 2023; Stella et al., 2023). While AI models are increasingly used to support personalised learning and cognitive assessment (Burton et al., 2024), there is still much to learn about their ability to replicate human cognitive and emotional structures in educational contexts (Stella et al., 2023).

**Manuscript aims, research questions and outline**

In this study, we use behavioural forma mentis networks (rather than textual forma mentis networks, as discussed by Haim et al., 2025) to explore the mindsets of different groups, focusing on PhD students and experts in STEM fields. We further subdivide the expert group into academics (research professors) and industrial professionals (experts working outside of academia). This allows us to investigate how individuals with various levels of expertise and professional experience structure their own mindset, i.e. ways of associating and perceiving ideas related to STEM subjects. Furthermore, we use GPT-3.5 alongside human participants to investigate how AI´s cognitive structures align with those of trainees, academics, and industrial experts, thus obtaining a baseline for discovering potential biases and limitations in both human and AI-generated associative networks (Stella et al., 2023).

Our study has two goals. Firstly, it aims to explore the cognitive and emotional associations that students, experts and academics form with concepts in STEM (Science, Technology, Engineering, and Mathematics). To achieve this, a behavioural forma mentis task was used, i.e. a continuous free word association task followed by a valence rating task. In the free association part, participants were presented with a series of STEM-related cue words and asked to respond with the first word or concept that came to mind. This method allowed for the mapping of associative networks that reflect the participants' cognitive structures and emotional perceptions of these disciplines. Secondly, this study also aims to compare the cognitive structures and emotional associations of STEM concepts between human participants and an AI language model, GPT-3.5. This comparison can identify potential biases in the AI's associative responses and assess its ability to replicate human-like cognitive patterns.

## 2. Methodology

### 2.1 Participants

The data collection took place online through a Shiny R application. For the human participants, a total of 209 individuals took part in the study. However, participants who left more than 25% of the responses blank, excluding the valence scores, were excluded from the final analysis. This left a total of 177 valid participants whose responses were used in the study. They were categorised into three groups: *trainees* (students at various levels, including PhD students), *experts* (professionals working outside of academia) and *academics* (those with a PhD degree and/or working at a university or as teachers). For the final analysis, 59 responses from trainees, 57 from experts, and 61 from academics were retained (52% female), representing over 25 different nationalities.

This study received the ethics approval from Warwick University (Ethical Application Reference: 09/17-18 approved by the Humanities and Social Sciences Research Ethics Sub-Committee of Warwick University).

For the GPT dataset, the same group distribution was replicated with GPT imitating 59 trainees, 57 experts and 61 academics. However, GPT-3.5 was not tasked with the imitation of different genders or nationalities but only with the imitation of these three groups. The full prompt administered to GPT-3.5 can be found in the appendix.

### 2.2 Behavioural Forma Mentis Networks

Both human participants and GPT-3.5 were presented with a set of 10 STEM-related cue words (art, biology, chemistry, complex, life, mathematics, physics, school, system, university). For each cue, participants were asked to provide three words that first came to mind, a process known as the associative response. This task is referred to as a continued free association task (De Deyne et al., 2013) because participants generate more than one response, allowing for a richer exploration of their cognitive associations. Research has shown that this method results in higher-quality data and a network that covers a broader segment of the human lexicon-semantic system, compared to tasks that limit participants to a single response (De Deyne et al., 2013).

After completing the word association task, participants were asked to rate the emotional valence of both the cue words and their associations using a Likert scale ranging from 1 (very negative) to 5 (very positive). This provided two layers of data: the semantic associations generated by the participants and their corresponding emotional responses. A neutral perception could be indicated by assigning a value of three, or, alternatively, by leaving a blank space if no strong sentiment was attached to the word.

The collected associations were used to create behavioural forma mentis networks (BFMNs), representing the cognitive and emotional landscapes of participants across the three groups. To construct the BFMNs, we linked each cue word to all its associated responses provided by the participants, receiving a directed network. The directed links represent the directionality of each cue word to its three associations, which makes it distinct from fluency networks. The nodes in the network represent both the cue words and their responses, with the links between forming the network edges. To obtain group-level BFMNs, we merged associations contributed by individuals within the same group of expertise (trainees, experts, academics). This allowed us to create networks that reflected the shared cognitive and emotional structures of each group of expertise.

We integrated the valence scores (on a scale from 1 to 5) obtained from the participants into the networks, adding an emotional layer to the network. Each node was assigned a valence label (positive, neutral or negative) based on statistical analysis of its valence scores using the Kruskall-Wallis test (with a significance threshold of $p < 0.1$). The Kruskall-Wallis test was used to compare the median ranks of valence scores of a given word ($w_i$) to the rank distribution of all other valence scores ($S_{j \neq i}$) in the dataset. Words with significantly higher median rank scores than the rest of the distribution were labelled as positive, while words with significantly lower valence scores labelled as negative. If no significant difference was detected, the word was labelled as neutral (Stella et al., 2019).

To visualise these valence labels within the BFMNs, each node was colour-coded: blue for positive, red for negative, and grey for neutral. Edges connecting nodes were also coloured accordingly based on the valence of the nodes they connect. Edges between two positive nodes were marked in blue, and in red for two negative nodes. Edges connecting at least one neutral node with either a positive or negative node are coloured in grey. Lastly, edges connecting positive and negative nodes were highlighted in purple to indicate contrasting emotional perceptions. Figure 1 (left) below shows an example of a forma mentis network created for the entire group of GPT-3.5 impersonating science academics.

## 2.3 Semantic Frames

Another way to assess the structure of behavioural forma mentis networks is via semantic frames (Stella et al., 2019). While a whole BFMN consists of the entire dataset of cue words and associations provided by the participants, semantic frames only consider the associations given to a specific target word. In more detail, semantic frames are network neighbourhoods of individual words within the topology defined by a behavioural forma mentis network. Figure 1 (right) shows an example of semantic frames for the cue words "art" and "mathematics" provided by GPT-3.5 academics in their BFMN. In the semantic frames, cue words are depicted in bigger size, with links only present between the cue and its associations but no links between the associations themselves. Since semantic frames are subgraphs of forma mentis networks, nodes and links are colour-coded to indicate positive (blue), negative (red) or neutral (grey) valence assigned by the participants. On the left, Figure 1 also shows via yellow arrows where in the forma mentis network the semantic frames for "art" and "mathematics" are located.

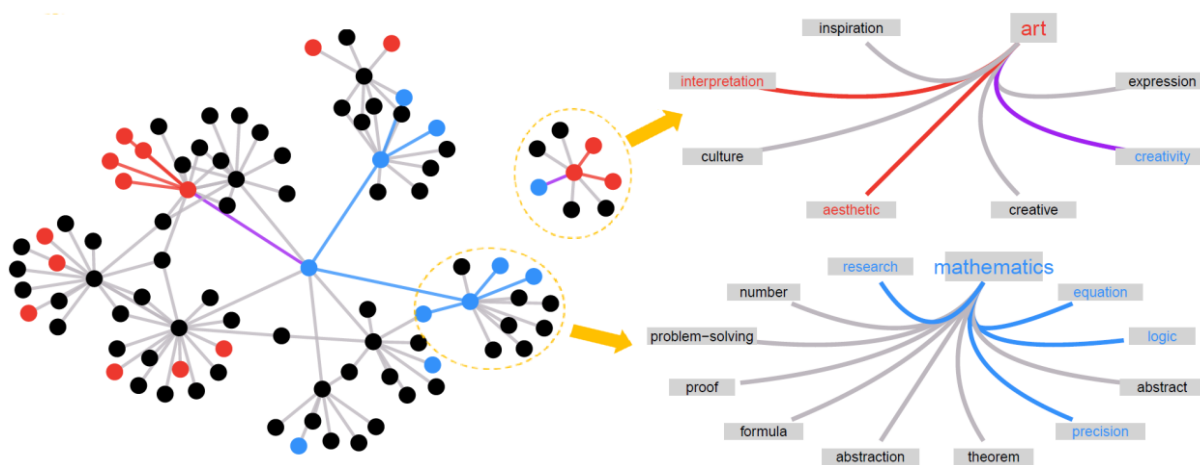

*Figure 1*. Example of a forma mentis network and its semantic frames from GPT-3.5 impersonating science academics. Left: Entire group-based network for GPT-3.5 academics for all cue words taken together. Right: Semantic frames for the cue words "art" and "mathematics" from GPT-3.5 academics. The semantic frame for "art" can be seen as a disconnected graph in the entire group-based network, as indicated by a yellow arrow.

Thus, semantic frames encapsulate the complex web of associations a person forms with a word and how these are positively/negatively perceived (Stella, 2022b). Semantic frames are based on frame semantics, a theory in computational linguistics and cognitive science according to which the meaning of a given word can be reconstructed by looking at its associated concepts (Fillmore & Baker, 2001). A complete list of the semantic frames for each cue word and each group can be found in the Appendix.

**2.4 Data Preprocessing**

The preprocessing of the data involved lemmatization, i.e. converting all responses to their base form using spaCy's lemmatizer (https://spacy.io, Last Access: 21/01/2025) to ensure consistency in word forms. Furthermore, all words were converted to lowercase, and spelling checks were conducted to correct any errors. Additionally, plural forms were standardised to their singular counterparts to maintain consistency across the dataset. Single-letter words or incomprehensible words were removed. Some human participants provided incomplete cue sets, i.e. only providing two instead of three associations for a given cue word. However, these responses were not discarded as the variation in the number of cue-word links was minimal and did not significantly impact the overall results. To ensure the network focused on robust associations, we filtered out idiosyncratic responses, by only retaining associations that were mentioned by at least two participants.

**2.5 Network Measures**

As discussed already, behavioural forma mentis networks are built from direct cue-response associations rather than fluency-based networks. To analyse the structural properties of group-level BFMNs, we considered four common network measures: average shortest path length (ASPL), diameter, modularity, and clustering coefficient (CC). These measures have been widely used in cognitive network research to assess the organization and connectivity of conceptual representations (Haim & Stella, 2023; Siew et al., 2019b).

ASPL quantifies the average number of steps required to travel between nodes in a network, capturing how closely interconnected concepts are and reflects the efficiency of information transfer. This measure has been linked to cognitive processing, with longer path lengths associated with slower reaction times and increased perceived dissimilarity between concepts (Kenett et al., 2018; Newman, 2010).

In contrast, network diameter, another global measure of distance, represents the longest shortest path between any two nodes in the network. While ASPL reflects the average number of steps it takes to travel the shortest distance between any two given nodes, diameter captures the furthest distance through the entire network. Thus, it provides insights into the spread and overall reachability of concepts (Newman, 2010; Haim et al., 2025).

Modularity is a mesoscopic measure and captures the extent to which a network can be decomposed into distinct subcommunities, with higher modularity indicating a more fragmented structure. Studies suggest that high modularity may reduce cognitive flexibility and hinder creative problem-solving (Siew et al., 2019b).

Finally, clustering coefficient (CC) describes how interconnected the neighbours of a node are. Higher CC values indicate stronger local connectivity, which has been linked to faster word recognition, improved memory retrieval, and enhanced learning (Newman, 2010; Siew et al., 2019b). CC is a crucial measure in cognitive network science as it quantifies the degree to which concepts in a semantic network are locally interconnected. Unlike global measures such as average shortest path length, network diameter or modularity - which capture overall network organisation - CC directly reflects the fine-grained, local connectivity among concepts. This local structure has been linked to cognitive processes including word recognition, semantic fluency, memory recall, and learning (Siew et al., 2019b; Siew, 2020). In the context of learning or expertise, the cognitive networks of experts have been shown to be more clustered with better-defined hierarchies and triadic closures than those of novices (Koponen & Pehkonen, 2010). Higher clustering is advantageous because it naturally increases the number of associations among nodes. As expertise develops through the reinforcement of conceptual links, this leads to more efficient knowledge integration and retrieval in the networks of experts. These findings have also been observed in concept maps (Koponen & Pehkonen, 2010; Thurn et al., 2020).

Given our focus on how concepts are organised at a micro-level, we prioritise CC in our analysis while briefly touching on broader network properties like ASPL, diameter and modularity in the analysis.

The clustering coefficient is a key measure of how tightly interconnected the nodes in a network are. Specifically, it evaluates the extent to which the neighbours of a given node are also connected to each other, forming triads or clusters (Koponen & Pehkonen, 2010; Siew et al., 2019b). For a given node $i$, its neighbourhood $\partial_i$ includes the nodes directly connected to $i$. The local clustering coefficient $C_i$ is defined as the ratio of the number of connections between neighbours of $i$ to the total number of possible connections those neighbours could have (Haim & Stella, 2023). Mathematically, it is expressed as:

$$C_i = \frac{|\{(k,l) \epsilon\, E \; for \; k, l \; \epsilon \; \partial_i\}|}{\frac{|\partial_i|(|\partial_i| - 1)}{2}}$$

where $E$ represents the set of edges in the network. The value of $C_i$ ranges from 0 (none of the neighbours of $i$ are connected) to 1 (all neighbours of $i$ are fully interconnected), see also Haim and Stella (2023).

In this study, we utilised the mean clustering coefficient to compare the structural complexity of behavioural forma mentis networks (BFMNs) generated by human participants (trainees, experts and academics) and GPT-3.5 (imitating trainees, experts and academics). The mean clustering coefficient is calculated as the average of the local clustering coefficients across all nodes in the network (Haim & Stella, 2023). To calculate the clustering coefficients, we constructed networks for each group, linking nodes based on participants' associative responses. This allowed us to measure how concepts related to one cue word (e.g. mathematics) were also associated with others (e.g. physics), revealing patterns of conceptual overlap and memory recall across different groups. The overall or mean clustering coefficient of the network is derived as the average of the local clustering coefficients for all nodes (Siew et al., 2019b).

## 3. Results

In our comparison across the three participant groups and between human and GPT data, we identified three major findings: (i) global-level network differences in the overall structure of behavioural forma mentis networks, (ii) local-level variations in semantic frames and (ii) differences in clustering patterns.

### 3.1 Global level differences in forma mentis networks

We created the full BFMNs for each group (human vs. GPT) and proficiency level (trainee, expert, academic) and calculated the following network measures: average shortest path length (ASPL), diameter, clustering coefficient (CC), and modularity. To assess whether the observed network measures were significantly different from what could be expected by random chance, we performed a reshuffling experiment. We generated 500 configuration models, i.e. random graphs with the same degree sequence as the original networks (Newman, 2018). We then compared network features of these random graphs to the empirical values, i.e. the actual values for ASPL, diameter, CC and modularity (see Table 1). Table 1 below presents the empirical values alongside the mean random values obtained from the 500 random graphs.

|  | Average Shortest Path | | | Diameter | | |
| --- | --- | --- | --- | --- | --- | --- |
|  | Empirical | Conf. Model (Mean) | p-value | Empirical | Conf. Model (Mean) | p-value |
| **Human Trainee** | 3.06 | 2.63 | .332 | 7.00 | 5.73 | .436 |
| **Human Expert** | 2.15 | 2.09 | .854 | 5.00 | 4.32 | .554 |
| **Human Academic** | 2.04 | 2.12 | .796 | 4.00 | 4.41 | .999 |
| **GPT Trainee** | 1.40 | 1.60 | .39 | 3.00 | 3.43 | .999 |
| **GPT Expert** | 1.19 | 1.28 | .456 | 2.00 | 2.48 | .999 |
| **GPT Academic** | 1.24 | 1.44 | .212 | 2.00 | 2.92 | .458 |

|  | Clustering Coefficient | | | Modularity | | |
|---|---|---|---|---|---|---|
|  | Empirical | Conf. Model (Mean) | p-value | Empirical | Conf. Model (Mean) | p-value |
| **Human Trainee** | 0.57 | 0.23 | < .001 | 0.66 | 0.65 | .786 |
| **Human Expert** | 0.36 | 0.18 | .002 | 0.69 | 0.68 | .210 |
| **Human Academic** | 0.42 | 0.19 | < .001 | 0.69 | 0.67 | .026 |
| **GPT Trainee** | 0.27 | 0.07 | < .001 | 0.72 | 0.68 | < .001 |
| **GPT Expert** | 0.17 | 0.04 | .004 | 0.73 | 0.71 | .112 |
| **GPT Academic** | 0.26 | 0.06 | .002 | 0.72 | 0.67 | .002 |

*Table 1*. *Comparison of network measures obtained from the actual graphs of human and GPT participants (empirical values) and 500 random graphs (mean random values) across different proficiency levels (trainee, expert, academic).*

As visible from Table 1, in contrast to human networks, GPT-3.5 consistently produced shorter ASPL, smaller network diameters and lower CC across all proficiency levels. Finally, modularity showed minimal differences between GPT and human networks. When comparing the empirical networks to the random graphs, only the measure of clustering coefficient consistently showed significant differences across all participant groups. No significant differences were observed between empirical and random graphs for ASPL and diameter, with modularity showing significant differences in three out of six cases. Thus, clustering coefficient emerges as the most important of the four measures to differentiate between empirical and random networks, which supports our focus on CC in the remaining analysis.

**3.2 Semantic Frames and Emotional Perceptions**

After the general-level comparison across the forma mentis networks generated by human participants or GPT-3.5, we investigate how each group perceived STEM-related concepts. Across all cue words, behavioural forma mentis networks (BFMNs) were created for each category (trainees, experts, and academics) both for human participants and for GPT-3.5 (see Figure 2). These networks offer a visual and structural representation of how different concepts are interconnected through participants' associations. The BFMNs of human participants (see Figure 2, left column) displayed dense, well-connected structures (average measures across groups: ASPL = 2,42; Diameter = 5,3; CC = 0,45; Modularity = 0,68). All human-generated networks exhibited strong interconnectivity between cue words, indicating robust associative patterns where multiple cue words were frequently linked together. This reflects a complex and integrated cognitive representation of STEM topics, with no isolated clusters.

Conversely, the behavioural forma mentis networks of GPT-3.5 (Figure 2, right column) were less dense and more sparse compared to those of human participants (average measures across groups: ASPL = 1,27; Diameter = 2,3; CC = 0,23; Modularity = 0,72). Compared to human networks, GPT-3.5 produced networks with shorter diameter, lower clustering and higher modularity. This indicates that human networks span a broader conceptual space with more robust integrated associative structures. GPT-3.5´s responses were more constrained with reduced integration of concepts into the overall structure. In several cases, key cue words such as "chemistry" in the trainee and expert categories, and "art" across all categories, were isolated from the rest of the network. These isolated words were not connected to other cue words, indicating that GPT-3.5's associative responses did not possess the same level of interconnectedness seen in human networks. This fragmentation suggests a gap in GPT-3.5's ability to form broad, coherent conceptual maps similar to human cognition, particularly when it comes to interdisciplinary connections like those between STEM and non-STEM fields.

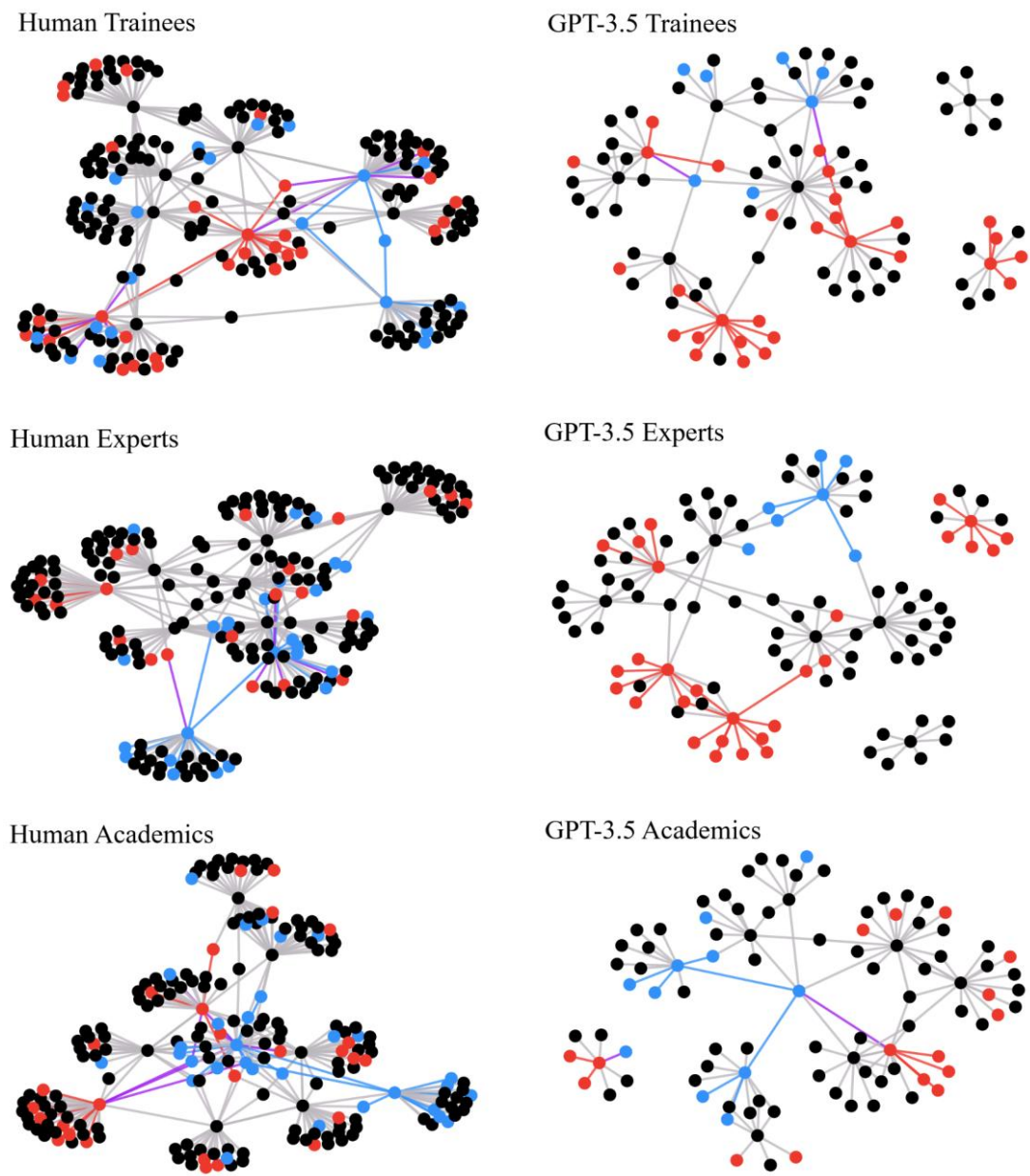

*Figure 2.* Group-based behavioural forma mentis networks across all stimuli and responses by humans (left) and GPT-3.5 (right) separated by level of expertise. Notice that only the networks for the GPT-3.5 impersonations show some disconnected graphs.

Overall, these results underscore that while GPT-3.5 can generate meaningful associations and semantic frames within STEM domains, it struggles with creating the same rich, interconnected networks that humans naturally produce (see Table 1 for a comparison of network measures). Furthermore, the AI's perception appears more vulnerable to biases introduced by task framing, as evidenced by its unexpectedly negative perception of art when primed to focus on science.

The further analysis of separate semantic frames provides deeper insights into the associative and emotional structures surrounding specific cue words such as "mathematics", "physics", "science", and "art" across different participant groups (trainees, experts, and academics). In our study, we observed that human participants consistently produced positive or neutral associations for STEM-related concepts, such as for "biology" and "mathematics" (Fig. 3). This finding diverges from past work, such as Ciringione et al. (2024), which linked mathematics to anxiety and negative emotional perceptions. In contrast, the participants in our study, across all groups, generally displayed neutral or positive emotional responses to these subjects, highlighting a potential shift in the way these topics are cognitively framed by different groups.

GPT-3.5, when tasked with the same word association exercise, demonstrated similar patterns of positive associations with STEM subjects but with noticeable differences. Most notably, when prompted to act as a "science expert", GPT-3.5 shifted its emotional and associative responses toward a negative perception of "art" (Fig. 4).

*Figure 3.* Semantic frames for the cue word "mathematics" by humans (left) and GPT-3.5 (right) separated by level of expertise (trainees, experts, academics).

*Figure 4.* Semantic frames for the cue word "art" by humans (left) and GPT-3.5 (right) separated by level of expertise (trainees, experts, academics).

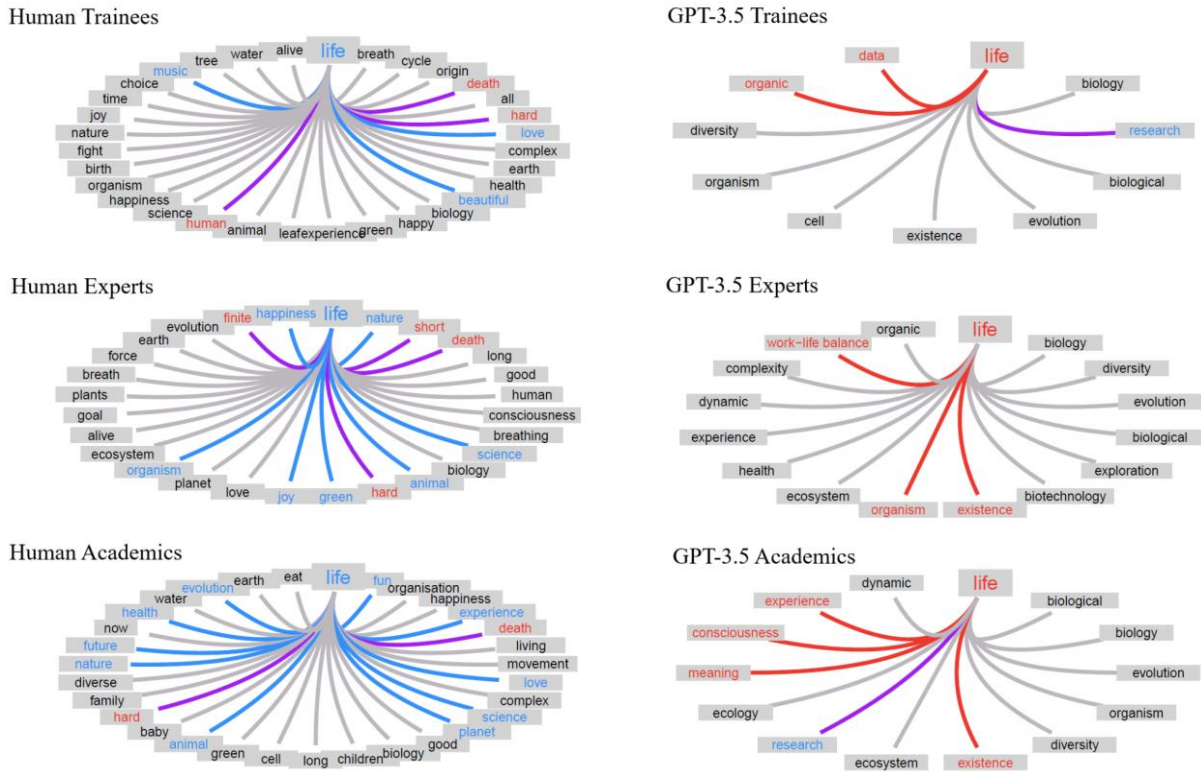

*Figure 5.* Semantic frames for the cue word "life" by humans (left) and GPT-3.5 (right) separated by level of expertise (trainees, experts, academics).

These findings highlight the nuanced differences between human and AI-generated semantic frames, particularly in how interdisciplinary concepts such as art and science are perceived. While human participants demonstrated a more balanced and integrated view, GPT's associative networks were narrower, shaped by the specific constraints of the task. This divergence underscores the importance of considering contextual prompts and task framing when using AI models like GPT in cognitive and semantic analysis.

**3.2 Clustering Coefficient Simulations**

Our analysis revealed further that human forma mentis networks exhibited substantially higher clustering than those generated by GPT-3.5 across all participant categories (see Table 2). This difference in clustering indicates that, in their mindsets, human participants tend to form more tightly interconnected associations across multiple cue words, reflecting more robust conceptual networks. These results align with the idea that higher clustering is associated with greater cognitive competence and advanced knowledge (Siew et al., 2019b). For example, a high clustering coefficient for an individual suggests that their knowledge is organised into cohesive groups of related concepts, facilitating efficient retrieval and integration of information. A higher clustering coefficient in human networks than in networks obtained from GPT-3.5 data indicates that humans are more adept at integrating diverse STEM concepts into coherent mental structures (Siew, 2020). In contrast, GPT-3.5, while partially replicating human clustering patterns, particularly in the lowered clustering coefficient observed in the expert category, demonstrated weaker conceptual integration, suggesting that AI-generated networks are less cohesively structured than those of human participants (Abramski et al., 2023).

|  | **Human** | **GPT-3.5** |
|---|---|---|
| **Trainee** | 0.572 | 0.273 |
| **Expert** | 0.358 | 0.172 |
| **Academic** | 0.417 | 0.257 |

*Table 2.* Clustering coefficients for cognitive networks constructed across all ten cue words and their responses provided by human participants and GPT-3.5.

To assess whether the observed clustering coefficients were significantly different from what could be expected by random chance, we again performed a reshuffling experiment (see Section 3.1). We generated 500 random graphs with the same degree sequence as the original networks and compared the clustering coefficients of these random graphs to the empirical values, i.e. the actual clustering coefficients observed in our data (see Figure 6).

For the human networks (Figure 6, left column), the empirical average clustering coefficients (represented by the red dashed lines in Figure 6) were considerably higher than the randomly generated values across all groups (p-values < 0.001 for all human groups). For GPT-3.5 (Figure 6, right column), the empirical clustering coefficients were generally closer to those of the random graphs, suggesting that the clustering observed in its networks could more likely be attributed to random variation (p-values < 0.01 for GPT-trainees and GPT-academics; p-value > 0.05 for GPT-experts). These findings highlight that, while GPT-3.5 can partially reproduce human clustering patterns, its networks are less structured overall in contrast to the networks of human individuals. Consequently, GPT-3.5 falls short in producing the intricate conceptual networks characteristic of human cognition.

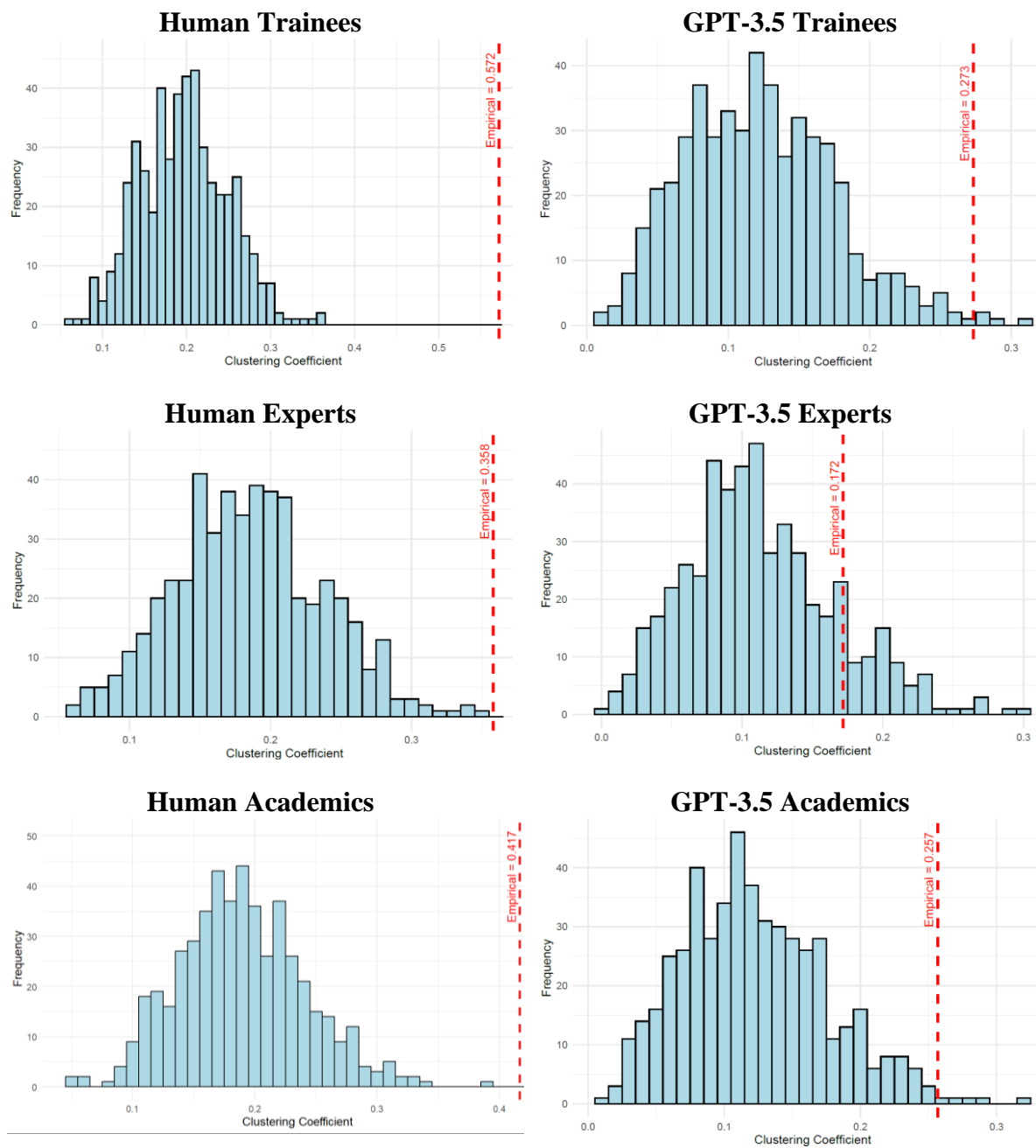

*Figure 6.* Distribution of clustering coefficients by humans (left column) and GPT-3.5 (right column) separated by level of expertise.

The higher clustering coefficients observed in human networks, particularly among experts, suggest a stronger presence of triadic closure, a principle where if two individuals (or nodes) are both connected to a third, they are more likely to connect to each other, forming a triad (Opsahl, 2013). This indicates a more integrated and advanced knowledge structure, where information is more readily accessible through multiple interconnected pathways. In contrast, GPT-3.5's weaker clustering, particularly among trainees and academics, points to a less cohesive network, where associations are less likely to form triads. This suggests that while GPT-3.5 can mimic some elements of expert knowledge, its ability to form complex, interconnected mental structures matched with expert knowledge is still limited compared to humans (Jo et al., 2024; Kozachek, 2023).

## Discussion

This study explored the cognitive and emotional associations that humans and GPT-3.5 form with STEM concepts, revealing important insights into the strengths and limitations of GPT-3.5 in replicating human-like cognitive structures. Our findings highlight significant differences in the patterns of clustering and complexity in the cognitive networks between human participants (trainees, experts and academics) and GPT-3.5. Human networks demonstrated more tightly interconnected knowledge structures with significantly higher clustering coefficients, reflecting greater competence and a better integration of STEM concepts into their cognitive networks. In contrast, GPT-3.5 produced less cohesive structures with lower clustering coefficients, indicating a weaker ability to form triadic closures and complex conceptual structures in the network.

These findings align with recent research showing that LLM-generated semantic networks exhibit poorer interconnectivity, lower local association organisation, and reduced flexibility compared to human networks (Wang et al., 2025). LLMs might engage in a more rigid, less efficient exploration of semantic space, which limits their ability to mimic human-like conceptual integration (Wang et al., 2025).

Higher clustering coefficients suggest the presence of stronger triadic closure, a process where triplets of nodes tend to be all linked with each other (Newman, 2018). For instance, consider "bed" linked to "mirror" and "table" linked to "mirror". In presence of triadic closure, these 3 nodes would all be linked with each other (i.e. form a complete graph) in case there was a link also between "bed" and "table". In cognitive network science, triadic closure might correspond to a facilitative effect for achieving more efficient information retrieval and stronger connections between concepts (Opsahl, 2013). For instance, in our example triadic closure would make it possible to access "bed" not only from "mirror" but also from "table", given the links between these 3 nodes. Thus, higher clustering in human networks is indicative of robust integration and advanced organisation of STEM-related knowledge, particularly among experts (Siew et al., 2019b). In highly clustered semantic networks, the presence of denser connections among neighboring nodes can enable the rapid retrieval of closely related concepts, which can facilitate the refinement of and elaboration of ideas (Siew et al., 2019b). Such structures may be particularly beneficial in the later stages of the creative process, where individuals must integrate disparate elements into cohesive and innovative solutions (Mednick, 1962). For instance, high clustering can support associative fluency within a well-defined domain since triadic closures foster the generation of novel combinations of distantly related concepts (Haim et al., 2024b; Kenett et al., 2014). Additionally, strong clustering can enhance the richness of idea exploration from semantic memory. This is especially critical for creativity in STEM fields, where innovative solutions often emerge from building on existing knowledge structures (Siew et al., 2019a).

Furthermore, the high clustering observed in human cognitive networks aligns with past findings where clustering emerges as a fundamental feature of language learning. In the mental lexicon, knowledge gaps emerge as learners first acquire broad conceptual landmarks that establish a foundational structure. For example, when learning about fruits, a child might first learn the words "apple" and "orange", leaving a big gap of possible other fruits connecting both to "apples" and to "oranges". Over time, these gaps are gradually filled with more specific and detailed information (e.g. "kiwi"), refining the overall network and increasing its clustering (Hills & Siew, 2018; Sizemore et al., 2018). This dynamics supports efficient information organisation and retrieval. Additionally, curiosity-driven exploration, especially deprivation curiosity, of the mental lexicon plays a key role in shaping these clustered structures. Deprivation curiosity is characterised by consciously seeking information in order to close knowledge gaps and a greater tendency to return to previously visited concepts. These curiosity-driven explorations promote the development of tightly clustered networks through uncertainty reduction (Lydon-Staley et al., 2021). These processes highlight the adaptability of human cognition to optimise knowledge representation through gap-filling and curiosity-driven clustering, which is lacking in GPT-3.5´s mental representation.

In contrast, GPT-3.5 demonstrated lower clustering coefficients and weaker triadic closure, indicative of a less interconnected knowledge structure. GPT-3.5´s clustering coefficients were closer to those obtained from random graphs and highlight GPT-3.5´s limitation in forming the intricate associative structures characteristic of human cognition ( Stella et al., 2019). This aligns with findings showing that open-source LLMs, such as GPT-models or BERT, excel in general feature extraction. However, they tend to fall short in capturing conceptual relationships reflecting the intricate structures typical for human cognitive networks (Aeschbach et al., 2024; Hussain et al., 2024). LLMs like GPT-3.5 can model certain aspects of free association in human individuals, such as capturing semantic similarity and generating plausible responses. However, the coherence and depth of their networks remain limited, as indicated by the lack of triadic closures characteristic for human cognition (Aeschbach et al., 2024; Hussain et al., 2024).

The second focus of the present study lies in examining emotional valences of STEM-related concepts. Our findings show that human participants across the three groups of expertise (trainees, experts, academics) generally displayed positive or neutral emotional responses to STEM concepts. Some negative associations were present for the cue words "chemistry", "school" and "system", diverging from past studies linking STEM fields to anxiety in students (Ciringione et al., 2024). STEM-unrelated words like "art" and "life" were consistently seen as positive by human participants. In comparison, GPT-3.5 exhibited more negative associations throughout the cue words. We found a striking difference between the valences assigned by humans and GPT-3.5 for the cue words "art" and "life", which were rated positively by human participants, but negatively by GPT-3.5. This may be influenced by the task framing. Since GPT-3.5 was prompted to respond as a science expert or science academics, this may have biased GPT-3.5 to respond neutrally to STEM concepts but negatively to STEM-unrelated concepts.

This possible context-dependence highlights the potential biases of AI models towards STEM fields, compared to the more balanced associations formed by human participants (Stella et al., 2023). These results contrast with prior findings from Abramski et al. (2023), where GPT-3.5 viewed "art" positively, suggesting that the prompting context plays a critical role in shaping GPT's semantic frames. The negative framing of art in our study could reflect an over-specialization or narrowing of GPT's responses due to the specific framing of the task, highlighting how an AI's perception can be more rigid and context-dependent compared to human cognition. This also demonstrates a limitation in GPT-3.5's ability to balance interdisciplinary associations, as human participants showed more nuanced and integrated associations between STEM and non-STEM fields.

**Limitations and future research**

While our study provides valuable insights into the cognitive and emotional associations formed by human participants and GPT-3.5 in STEM contexts, we need to address some limitations of this research.

One limitation lies in the sample size and demographic diversity of the human participants. Although our study included a range of trainees, experts and academics, the relatively small sample sizes (~ 60 participants in each group) may not properly represent these populations. Similarly, GPT-3.5´s simulations of these groups were not informed by additional demographic factors like age, gender or cultural background, which could limit GPT-3.5´s associative and emotional patterns during the simulations.

Furthermore, while the use of clustering coefficients and semantic frames provided meaningful insights, our analysis focused primarily on macro-level network metrics. Additional micro-level measures, such as node centrality or measures of distance, might reveal further nuances in associative structures.

Future research could focus on enhancing GPT-3.5´s capacity to form richer and more interconnected networks, by improving its ability to model complex relationships between concepts. Addressing context-dependence, reducing biases and refining emotional modelling are essential steps toward aligning GPT-3.5´s cognitive and emotional frameworks with those of humans. Advances in these areas could strengthen the application of AI in education and cognitive research.

**Conclusion**

This research contributes to our understanding of how humans with different levels of advanced expertise in STEM (i.e., PhD students, academics and industry experts) shape their mindsets towards STEM disciplines. In particular, we map the mindset of these groups of individuals in terms of how they associate and perceive ideas, adopting the cognitive network framework of behavioural forma mentis networks. We find that all these groups can provide higher-than-random clustering levels of concepts, with mostly positive or neutral perceptions of STEM fields and subjects. These perceptions differ greatly from the negative and more anxious perceptions mapped within STEM high-school students. Furthermore, we test here GPT-3.5 as a reference null model, building behavioural forma mentis networks out of simulated experts. Although GPT-3.5 can understand the linguistic instructions behind the task and provide cognitive and emotional patterns, we identify several areas where it falls short, including negative biased perceptions towards "art" which depend on context. The implications for GPT models in educational contexts suggest a need for further improvements in the alignment with human-like knowledge structures and emotional perceptions. Instead, the implications of this study for human data are mostly relative to behavioural forma mentis networks outlining interesting and non-random patterns of knowledge clustering within STEM experts, with relevance for future studies relating advanced domain knowledge with semantic memory and emotional perceptions.

**Appendix**

**I. Full prompt administered to GPT-3.5**

The prompt that is used to generate the data with GPT-3.5 in the online ChatGPT interface is presented below. The first sentence (highlighted in bold below) is changed for each category corresponding to the specific persona.

**Trainee:** "Impersonate either a PhD-student or a post-graduate student working in a STEM-related field."

**Experts:** "Impersonate a person with a PhD in a STEM related field and a position at a professor level (in academia) who has a research record, a lab, etc."

**Academics:** "Impersonate a person with a PhD in a STEM related field and several years of experience in a corporal setting outside of academia."

Full prompt for "Trainee"

**Impersonate either a PhD-student or a post-graduate student working in a STEM-related field.** For each of the 10 cue words, 'physics', 'life', 'university', 'system', 'complex', 'art', 'chemistry', 'mathematics', 'school' and 'biology', you will perform the following tasks. First, rate each cue word from 1 (= very negative) to 5 (= very positive). Next, for each cue word you will perform an association task. Give 3 words that come into your mind when you think of the cue word. After which, you rate each word according to how you perceive each word, ranging from 1 (= very negative) to 5 (=very positive). Give your answer as followed, "cue word"="rating"="association 1"="rating"="association 2"="rating"="association 3"="rating"

**II. Semantic frames**

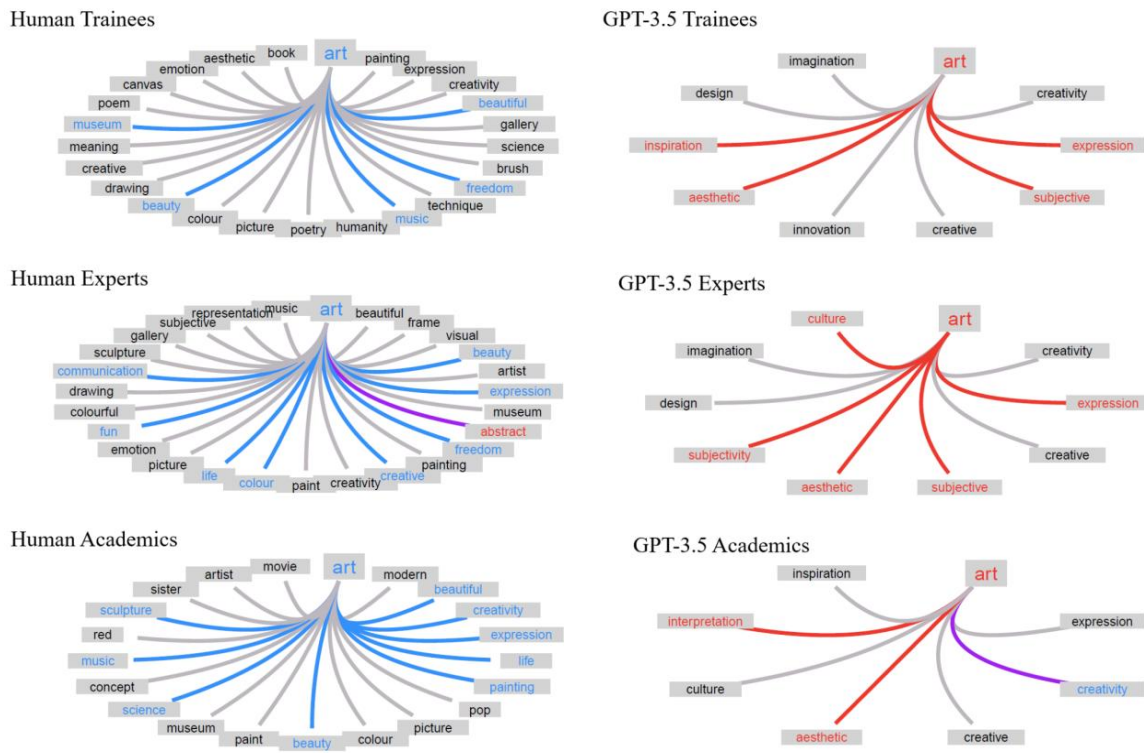

**Figure 7.** Semantic frames for the cue word "art" by humans (left) and GPT-3.5 (right) separated by level of expertise (trainees, experts, academics).

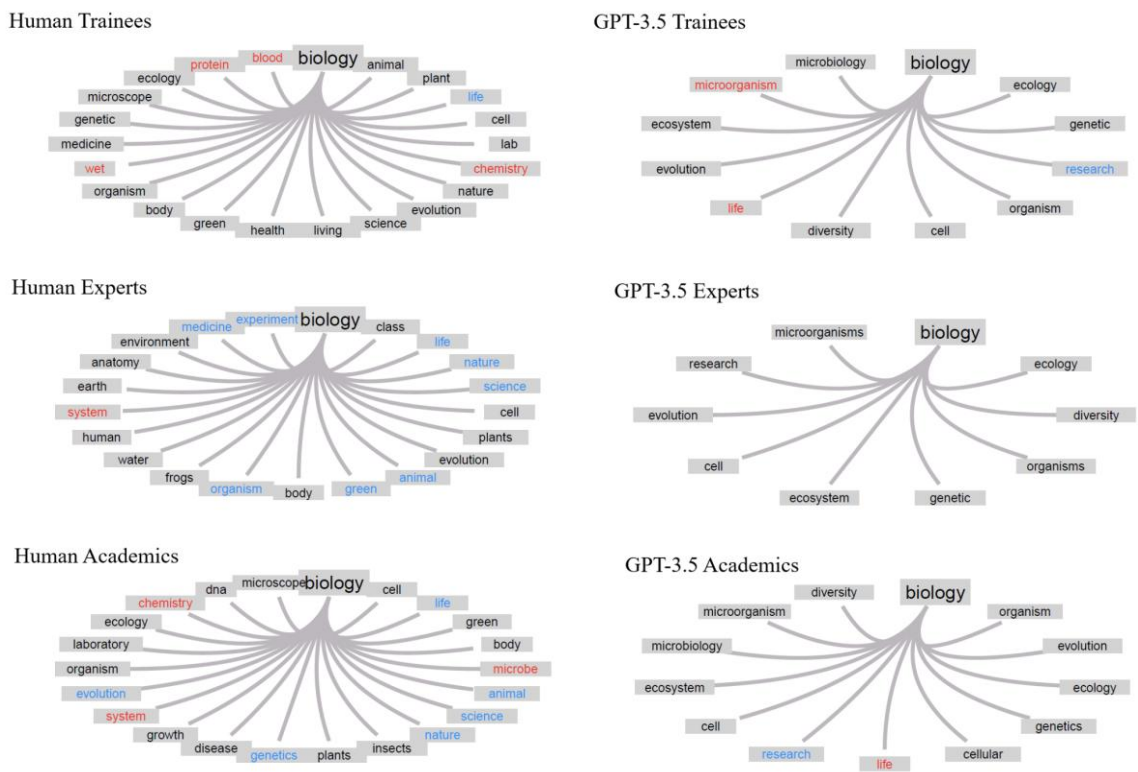

**Figure 8.** Semantic frames for the cue word "biology" by humans (left) and GPT-3.5 (right) separated by level of expertise (trainees, experts, academics).

**Figure 9.** Semantic frames for the cue word "chemistry" by humans (left) and GPT-3.5 (right) separated by level of expertise (trainees, experts, academics).

**Figure 10.** Semantic frames for the cue word "complex" by humans (left) and GPT-3.5 (right) separated by level of expertise (trainees, experts, academics).

**Figure 11.** Semantic frames for the cue word "life" by humans (left) and GPT-3.5 (right) separated by level of expertise (trainees, experts, academics).

**Figure 12.** Semantic frames for the cue word "mathematics" by humans (left) and GPT-3.5 (right) separated by level of expertise (trainees, experts, academics).

**Figure 13.** Semantic frames for the cue word "physics" by humans (left) and GPT-3.5 (right) separated by level of expertise (trainees, experts, academics).

**Figure 14.** Semantic frames for the cue word "school" by humans (left) and GPT-3.5 (right) separated by level of expertise (trainees, experts, academics).

**Figure 15.** Semantic frames for the cue word "system" by humans (left) and GPT-3.5 (right) separated by level of expertise (trainees, experts, academics).

**Figure 16.** Semantic frames for the cue word "university" by humans (left) and GPT-3.5 (right) separated by level of expertise (trainees, experts, academics).